\documentclass[10pt,twocolumn,letterpaper]{article}

\usepackage{cvpr}              
\usepackage[accsupp]{axessibility}

\usepackage[dvipsnames]{xcolor}

\definecolor{cvprblue}{rgb}{0.21,0.49,0.74}
\usepackage[pagebackref,breaklinks,colorlinks,citecolor=cvprblue]{hyperref}
\newcommand{\methodname}{USE}

\newcommand{\tb}[3]{\setlength{\tabcolsep}{#2mm}\begin{tabular}{#1}#3\end{tabular}}

\newcommand{\tablestyle}[2]{\setlength{\tabcolsep}{#1}\renewcommand{\arraystretch}{#2}\centering}

\def\paperID{16801} 
\def\confName{CVPR}
\def\confYear{2024}

\title{USE: Universal Segment Embeddings for Open-Vocabulary Image Segmentation}

\author{
\tb{@{}ccccc@{}}{3}{
Xiaoqi Wang$^{1, 2, 3}$&
Wenbin He$^{1, 2}$&
Xiwei Xuan$^{1, 2, 4}$&
Clint Sebastian$^2$&
Jorge Piazentin Ono$^{1, 2}$
}\\
\tb{@{}cccccc@{}}{3}{
Xin Li$^{1, 2}$&
Sima Behpour$^{1, 2}$&
Thang Doan$^{1, 2}$&
Liang Gou$^{1, 2}$&
Han-Wei Shen$^3$&
Liu Ren$^{1, 2}$
}\\
\tb{@{}cc@{}}{8}{
$^1$Bosch Research North America&
$^2$Bosch Center for Artificial Intelligence (BCAI)
}\\
\tb{@{}cc@{}}{8}{
$^3$The Ohio State University&
$^4$University of California Davis
}\\
{\tt\small wang.5502@osu.edu wenbin.he2@us.bosch.com xwxuan@ucdavis.edu, clint.sebastian@de.bosch.com}\\
{\tt\small \{jorge.piazentinono, xin.li9, sima.behpour, thang.doan, liang.gou\}@us.bosch.com}\\
{\tt\small shen.94@osu.edu liu.ren@us.bosch.com}
}

\begin{document}
\maketitle
\begin{abstract}

The open-vocabulary image segmentation task involves partitioning images into semantically meaningful segments and classifying them with flexible text-defined categories.  The recent vision-based foundation models such as the Segment Anything Model (SAM) have shown superior performance in generating class-agnostic image segments.  The main challenge in open-vocabulary image segmentation now lies in accurately classifying these segments into text-defined categories.  In this paper, we introduce the Universal Segment Embedding (\methodname) framework to address this challenge.  This framework is comprised of two key components: 1) \textbf{a data pipeline} designed to efficiently curate a large amount of segment-text pairs at various granularities, and 2) \textbf{a universal segment embedding model} that enables precise segment classification into a vast range of text-defined categories.  The USE model can not only help open-vocabulary image segmentation but also facilitate other downstream tasks (e.g., querying and ranking).  Through comprehensive experimental studies on semantic segmentation and part segmentation benchmarks, we demonstrate that the USE framework outperforms state-of-the-art open-vocabulary segmentation methods.

\end{abstract}

\section{Introduction}
\label{sec:intro}

Open-vocabulary image segmentation~\cite{simseg, openseg, ovseg, san} aims to partition images into semantically meaningful segments and classify them with arbitrary classes defined by texts.  Recent advances in vision foundation models such as the Segment Anything Model (SAM)~\cite{sam} have shown superior performance in grouping image pixels into semantically meaningful segments at various granularities (e.g., object, part, and subpart).  However, the existing open-vocabulary image segmentation methods~\cite{simseg, openseg, ovseg, san} face challenges in fully utilizing image segments generated by foundation models.  For instance, end-to-end methods such as side adapter network (SAN)~\cite{san} cannot take image segments generated by foundation models as input or prompt to assign class labels.  While two-stage methods (e.g., OVSeg~\cite{ovseg}) decouple the image segmentation and classification, they are still limited in classifying segments at various granularities because of limited human annotations~\cite{coco}.

\setlength{\abovecaptionskip}{6pt}
\setlength{\belowcaptionskip}{-16pt}
\begin{figure}[tb]
  \centering
  \includegraphics[width=\linewidth]{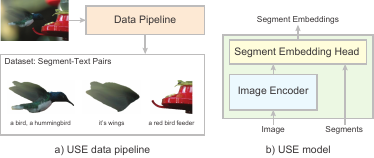}
  \caption{The proposed framework consists of two major components: a) data pipeline that generates segment-text pairs from image datasets and b) segment embedding model.}
  \label{fig:intro}
\end{figure}
\setlength{\abovecaptionskip}{10pt}
\setlength{\belowcaptionskip}{0pt}

In our study, we introduce a Universal Segment Embedding (USE) framework to tackle the identified challenges. The goal of USE is to take an image and various segments as input and generate an embedding vector for each segment that aligns with its corresponding text descriptions. These segment embeddings can then be utilized for classifying the segments in a zero-shot manner, similar to the CLIP~\cite{clip} model used for image classification.  Inspired by the recent advances in vision-language foundation models~\cite{kosmos2, llava}, we develop the USE framework with a data-centric approach.  Specifically, we introduce a \textbf{data pipeline} (Figure~\ref{fig:intro}a) designed to autonomously generate segment-text pairs at various granularities without human annotations.  In addition, we develop a lightweight universal segment embedding \textbf{model} (Figure~\ref{fig:intro}b) that can be trained efficiently on the large scale of segment-text pairs.

\textbf{Data Pipeline.}  Training data with a large scale of high-quality segment-text pairs plays an indispensable role in achieving a high-performing USE model.  Inspired by the foundation model-powered data-centric approaches~\cite{kosmos2}, we build a data pipeline that leverages a set of vision or vision-language foundation models to extract segment-text pairs from unlabeled images. Given an image, our data pipeline starts by generating detailed descriptions of the objects and parts in the image with a Multimodal Large Language Model (MLLM)~\cite{cogvlm}.  Then, we detect the most relevant bounding box for each object/part with a phrase grounding model~\cite{GDino}.  In the end, the segments of the objects and parts are generated based on the bounding boxes to collect segment-text pairs.

\textbf{Model.}  We develop the USE model by leveraging the capabilities of pre-trained foundation models with \textit{minimal trainable parameters}. The USE model consists of two major components, including an image encoder that is adapted from pre-trained vision foundation models and a lightweight segment embedding head that generates segment embeddings for input segments.  Note that the output of the image encoder can be reused with different segments, and the lightweight segment embedding head can generate embeddings efficiently.

We conducted extensive experiments on open-vocabulary semantic segmentation and part segmentation benchmarks to demonstrate the advances of the proposed data pipeline and model.  Our framework not only achieves state-of-the-art performance but also has flexibility in handling different open-vocabulary recognition tasks.

In summary, the contributions of this paper are threefold:
\begin{itemize}
  \item We propose a carefully designed data pipeline that can autonomously generate high-quality segment-text pairs at various granularities without human annotations.
  \item We propose a lightweight segment embedding model that can generate high-quality segment embeddings, which are well-aligned with text descriptions.  Hence, it enables various zero-shot image segmentation tasks such as semantic, instance, and part segmentation.  In addition, the embeddings offer efficient querying of image segments by text.
  \item Substantial performance improvements are observed with our approach over the state-of-the-art open-vocabulary image segmentation methods on different tasks including semantic and part segmentation.
\end{itemize}

\section{Related Work}
\label{sec:related-works}

\textbf{Multi-Modality Representation Learning.}  Recently, learning from large-scale image-text data (e.g., CLIP~\cite{clip}) has shown promising results in connecting visual concepts with textual descriptions.  Pre-trained CLIP~\cite{clip} has endowed many computer vision tasks with the capability of open-vocabulary recognition by learning a joint representation of image and text.  These computer vision tasks include but are not limited to image segmentation~\cite{ovseg, clips4, san, odise},  object detection~\cite{clip-od, regionclip}, and image captioning~\cite{clipcap}.  However, the multi-modality representation learning for segment-text data is still under-explored with very few existing work~\cite{ovseg, odise}.  OVSeg~\cite{ovseg} proposes a mask-adapted CLIP that fine-tunes CLIP on a collection of masked image regions to produce mask-aware image embeddings. Unfortunately, OVSeg fails to connect rich semantic information, such as object attributes, with the masked regions.  It also has the limitation that the background information outside the masked region is completely ignored during the generation of segment embeddings.  Unlike OVseg, the {\methodname} model can learn more expressive segment embeddings enriched with detailed text descriptions, including color, shape, size, etc.  In addition, the segment embeddings generated by the {\methodname} model will take the context information outside the masked region into account given the detailed text descriptions.

\textbf{Open-Vocabulary Image Segmentation.}  Driven by the increasing demands of real-world visual tasks, such as autonomous driving, the significance of open-vocabulary image segmentation is growing rapidly.  The existing methods can be classified into two categories: end-to-end approaches~\cite{groupvit,san,odise,x-decoder} and two-stage approaches~\cite{simseg,ovseg,ding2022open,huynh2022open}. The two-stage approaches first generate class-agnostic segment proposals and then classify segments into text-defined categories, whereas the end-to-end approaches often generate class-specific segments in an end-to-end manner.  Our approach aligns with the two-stage paradigm.  Compared with the previous two-stage methods, our approach can take segments of various granularities as input and generate the corresponding embeddings.  Meanwhile, we propose a foundation model-powered data pipeline to generate a large scale of segment-text pairs, which enhances the zero-shot ability of our model.

\textbf{Improving Image-Text Datasets.}  The careful curation of high-quality image-text pairs is the secret sauce behind the remarkable performance of large-scale pre-trained multimodal models like CLIP~\cite{clip}.  Inspired by this observation, researchers have recently conducted extensive research on improving the quality of image-text datasets, which can further improve the performance of open-vocabulary computer vision tasks.  The existing work can be categorized into two classes: data filtering~\cite{cao2023less, maini2023t} and data improvement~\cite{fan2023improving, zhu2023chatgpt, lai2023scarcity}.  Data filtering aims to improve the efficiency and robustness of model training by filtering noisy image-text pairs, while data improvement focuses on improving the alignment of image and text data.  In order to avoid filtering out images with rich visual concepts, we designed a data improvement approach as part of our data pipeline.  Similar to~\cite{zhu2023chatgpt}, we leverage MLLMs to infuse more informative visual concepts into image captions.  Furthermore, we propose to augment the image captions by meticulously describing the parts of objects in the image, thereby enriching the semantics of captions at multiple levels of granularity.

\section{Method}

In this paper, we propose a novel open-vocabulary image segmentation framework, {\methodname}, which consists of two key components: a data pipeline (Section~\ref{sec:method-data}) and a universal segment embedding model (Section~\ref{sec:method-model}).  Specifically, the data pipeline aims to automatically curate large-scale segment-text pairs with fine-grained object descriptions at multiple levels of granularity; the universal segment embedding model generates segment embeddings that are aligned with text embeddings in the joint space of vision and language.  Details of the two components are as follows.

\setlength{\abovecaptionskip}{6pt}
\setlength{\belowcaptionskip}{-16pt}
\begin{figure}[t]
  \centering
  \includegraphics[width=\columnwidth]{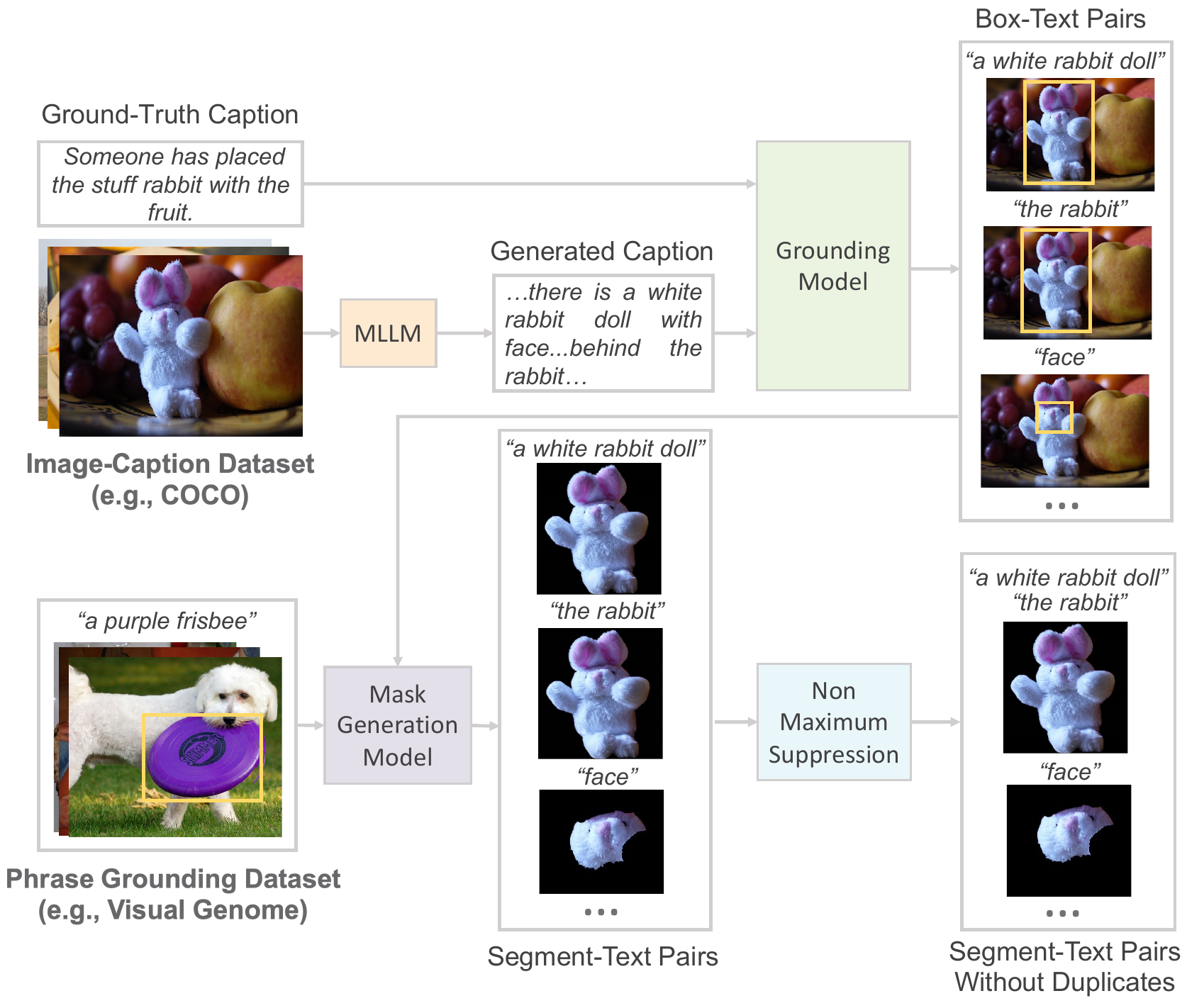}
  \caption{The overview of our data pipeline, which automatically constructs segment-text pairs at different levels of granularity.  We design a unified data pipeline that curates data from different types of data sources while taking advantage of multiple foundation models to streamline the process.}
  \label{fig:data-pipeline}
\end{figure}
\setlength{\abovecaptionskip}{10pt}
\setlength{\belowcaptionskip}{0pt}

\subsection{{\methodname} Data Pipeline}
\label{sec:method-data}

In this section, we introduce our data pipeline to automatically curate segment-text pairs whose semantics are closely aligned.  We carefully designed the data pipeline in a way that both the segments and text encapsulate information at multiple levels of granularity, with the purpose of enhancing the open-vocabulary recognition ability of our model.

The proposed data pipeline can be generalized for curating data from multiple types of data sources including image-only datasets (e.g., CIFAR-100~\cite{cifar100}), image-caption datasets (e.g., COCO~\cite{coco}, SBU~\cite{sbu}, and CC3M~\cite{cc3m}), and image with phrase grounding boxes (e.g., Visual Genome~\cite{visual-genome}).  This unified data pipeline consolidates the segment-text pairs extracted from different image datasets and generates a collection of segments for each image where each segment can have multiple text descriptions associated with it.  More importantly, this data pipeline is fully automatic and can be easily scaled up to billions of images.

The high-level overview of the proposed data pipeline is presented in Figure~\ref{fig:data-pipeline}.  It can be decomposed into three major modules: (a) an image captioning module that generates detailed descriptions of the image at different levels of granularity, (b) a referring expression grounding module that produces box-text pairs based on the images and captions, and (c) a mask generation module that converts box-text pairs into segment-text pairs.  A detailed illustration of our data pipeline is discussed in the following sections.

\setlength{\abovecaptionskip}{6pt}
\setlength{\belowcaptionskip}{-6pt}
\begin{figure}[ht]
  \centering
  \includegraphics[width=\columnwidth]{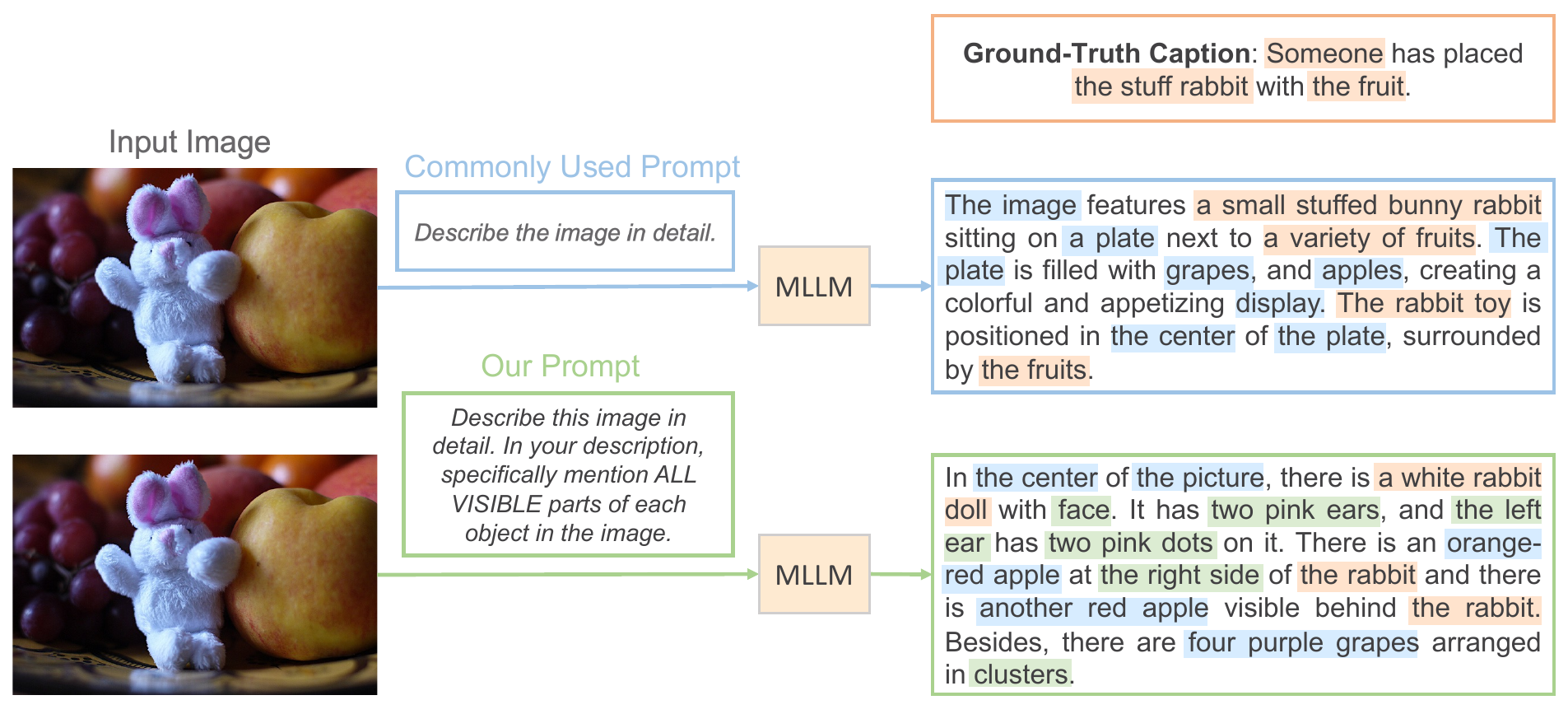}
  \caption{The examples of the ground-truth caption, the caption generated with the commonly used image captioning prompt, and the caption generated with our prompt. Our prompt can guide the MLLM to generate captions with more fine-grained object parts.}
  \label{fig:compare-caption}
\end{figure}
\setlength{\abovecaptionskip}{10pt}
\setlength{\belowcaptionskip}{0pt}

\textbf{Multi-Granularity Image Captioning.}  Our data pipeline starts with generating descriptions of objects (or parts) as well as their attributes from images.  The quality and diversity of the descriptions play an important role in extracting segment-text pairs that cover objects in images as much as possible.  We initially start with web-crawled or human-generated image captions (e.g., COCO~\cite{coco}, SBU~\cite{sbu}, CC3M~\cite{cc3m}) following previous work~\cite{clip, kosmos2}.  However, we observe that these captions either lack descriptions about object attributes or only focus on the main objects in the image (see the ground-truth caption in Figure~\ref{fig:compare-caption}).  This motivates us to generate image captions with richer semantic information.  To this end, we leverage the recent advances of MLLMs such as CogVLM~\cite{cogvlm}, Kosmos-2~\cite{kosmos2}, and LLaVA~\cite{llava}.  For all the MLLMs, the design of the text prompt is important for guiding the MLLMs to generate captions with desired properties.  In order to obtain detailed descriptions of objects and parts in images, we prompt the MLLMs as follows:

\textit{"Describe this image in detail. In your description, specifically mention ALL VISIBLE parts of each object in the image."}

Compared with the commonly used image captioning prompts (e.g., "Describe the image in detail."), our prompt allows MLLMs to not only describe the objects along with their attributes but also mention the visible parts of each object presented in the image.  As shown in Figure~\ref{fig:compare-caption}, the caption generated with our prompt specifically mentions "\textit{face}" and "\textit{two pink ears}" along with detailed descriptions of the color of the apple, while the caption generated with the commonly used prompt fails to include this level of fine-grained details about the image.  In our experimental study, we chose to employ CogVLM as the MLLM for generating multi-granularity captions.

\setlength{\abovecaptionskip}{6pt}
\setlength{\belowcaptionskip}{-6pt}
\begin{figure}[th]
  \centering
  \includegraphics[width=\columnwidth]{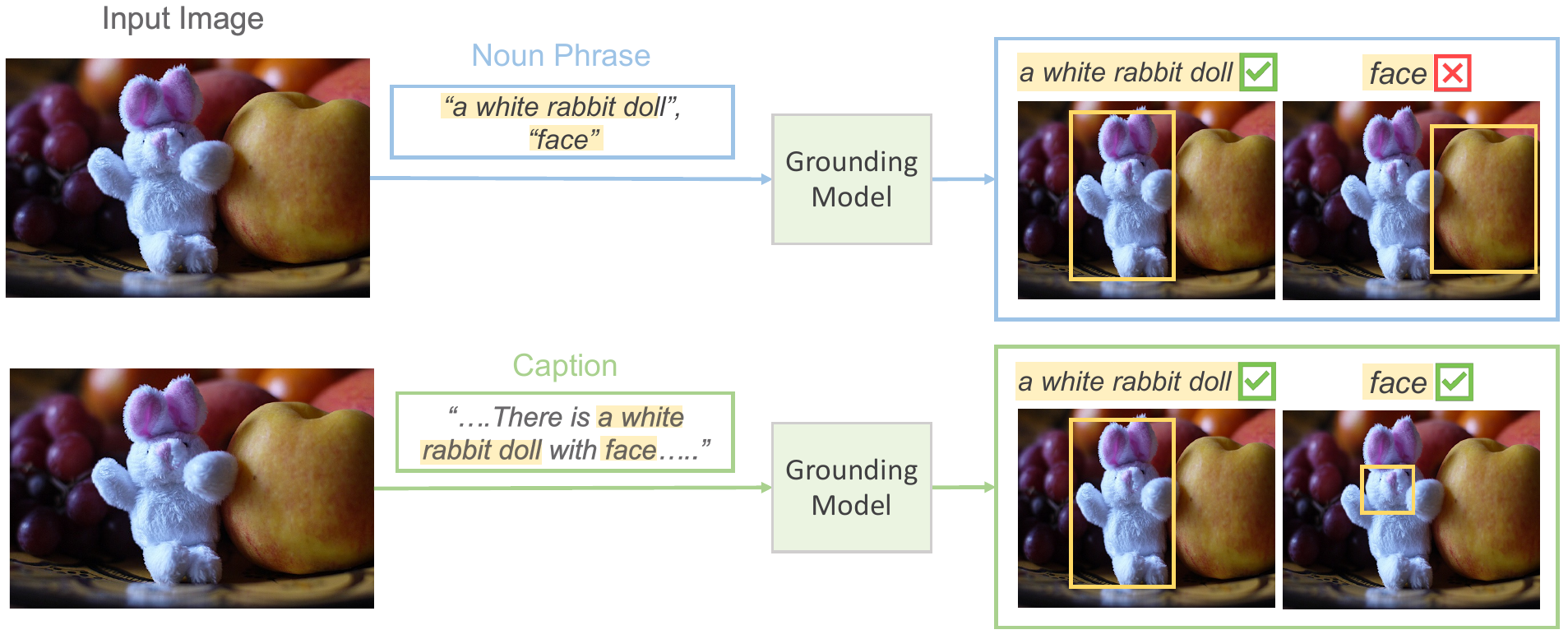}
  \caption{The examples of generated box-text pairs if we query the grounding model with either the entire caption or individual noun phrases.  Querying with the entire caption can help to accurately identify object parts by considering more context information.}
  \label{fig:grounding-query}
\end{figure}
\setlength{\abovecaptionskip}{10pt}
\setlength{\belowcaptionskip}{0pt}

\textbf{Referring Expression Grounding from Captions.}  Given the captions from different sources (i.e., ground-truth captions and MLLM-generated captions), the next step is extracting referring expressions from the captions and identifying their corresponding image regions represented by bounding boxes.  Inspired by Kosmos-2~\cite{kosmos2}, we first extract the noun phrases using spaCy~\cite{spacy} and then expand the noun phrases as referring expressions.  For example, from a caption ("\textit{There is an orange-red apple at the right side of the rabbit and there is another red apple visible behind the rabbit.}"), we can obtain the noun phrases (\textit{"an orange-red apple", "the right side", "the rabbit", "another red apple"}).  We further expand the noun phrases to referring expressions by recursively traversing the children of noun phrases in the dependency tree and concatenating them.  For the above example, the referring expressions we obtained after expanding noun phrases are (\textit{"an orange-red apple", "the right side of the rabbit", "the rabbit", "another red apple visible behind the rabbit"}). Clearly, referring expressions could capture more context information regarding the objects.  Existing open-vocabulary segmentation models that contain segment-text curation pipelines~\cite{ovseg, openseg} have a limited understanding of the text, either only including nouns (e.g., \textit{"apple", "side", "rabbit"}) from the caption, or including adjectives and nouns separately (e.g, \textit{"apple", "side", "rabbit", "orange-red", "red", "visible", "right"}). 

Compared with their approaches, the training data curated by our data pipeline will encapsulate richer semantics such that our open-vocabulary recognition ability can be enhanced and the predicted segments can be more consistent with the text query.

In order to obtain the bounding boxes associated with the extracted referring expressions, we adopt the open-vocabulary grounding models (e.g., Grounding DINO~\cite{GDino} and CoDet~\cite{CoDet}).  Note that some of the MLLMs~\cite{kosmos2} also offer the grounding capability, however, the generated bounding boxes are less accurate than the specialized grounding models.  In this work, we use the Grounding Dino as an example.  Given the image caption, there are two possible approaches to collecting bounding boxes associated with the noun phrases: querying with the noun phrases individually like what previous method~\cite{kosmos2} did or querying with the entire caption and then matching the boxes with the phrases.  We observe that querying with the entire caption allows the grounding model to capture the comprehensive referring relationships implicitly encapsulated in the caption.  In particular, when querying for object parts, the context is extremely important.  For example, as shown in Figure~\ref{fig:grounding-query}, the rabbit face can be accurately located when querying with the entire caption, while the face is mistakenly assigned with a bounding box containing the apple if we query with the noun phrase \textit{"face"} alone.  Hence, we decided to query the grounding model with the entire caption and match the boxes with the phrases as follows.  For each predicted box, we first identify the token with the highest probability score and associate the box with the noun phrase that contains the identified token.  As a result, we generate a collection of box-text pairs for the next step.  Note that we also extend box-phrase pairs to box-expression pairs and store both because the description of an image region can be ambiguous and from multiple levels of detail.

\setlength{\abovecaptionskip}{6pt}
\setlength{\belowcaptionskip}{-12pt}
\begin{figure*}[t]
  \centering
  \includegraphics[width=\linewidth]{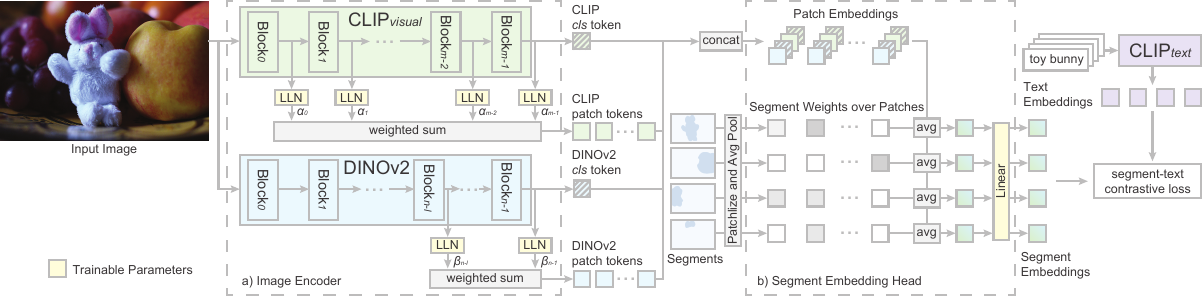}
  \caption{Architecture of the USE model, which consists of: a) an image encoder to extract image features for local patches and b) a segment embedding head maps the image features to segment embeddings that are aligned with text descriptions.  The USE model is trained with segment-text contrastive loss using the segment and text embeddings.}
  \label{fig:model}
\end{figure*}
\setlength{\abovecaptionskip}{10pt}
\setlength{\belowcaptionskip}{0pt}

\textbf{Mask Generation with Box Prompt.}  Given the box-text pairs generated by the referring expression grounding model mentioned above or directly from human annotations (e.g., Visual Genome~\cite{visual-genome}), the next step is to convert the bounding boxes into masks.  We employ the image segmentation model SAM~\cite{sam} which takes a bounding box as a prompt and outputs the mask of the best object that tightly fits with the box.  For each box, the SAM will generate multiple masks, and we only choose the one with the highest stability score (predicted by the SAM).  Similar to SAM, we perform two post-processing steps over the chosen masks including filling the small holes and removing the isolated small components.  We notice that for some text with vague meanings (e.g., a room, the atmosphere), the bounding boxes often cover the entire image.  If the size of the box is greater than 90\% of the image size, we directly use the mask of the entire image as the corresponding segments without using SAM.  Then, a collection of segment-text pairs can be obtained and merged via mask-based non-maximum-suppression (NMS).  We use NMS to remove duplicate masks for each image because different text descriptions may refer to the same object in the image.  After NMS, all the text descriptions associated with the duplicate masks will be merged and assigned to the corresponding mask.

\subsection{{\methodname} Model}
\label{sec:method-model}

Inspired by recent advancements in multi-modal foundation models (e.g., LLaVA~\cite{llava}, CogVLM~\cite{cogvlm}), we introduce the USE model, which leverages the capabilities of pre-trained foundation models (i.e., CLIP~\cite{clip} and DINOv2~\cite{dinov2}) with minimal trainable parameters. The architecture of the USE model is illustrated in Figure~\ref{fig:model}, comprising two major components: a) an image encoder that extracts image features by adapting the pre-trained foundation models, and b) a segment embedding head that generates segment embeddings based on the input segments and maps the segment embeddings to the vision-language space.  In the subsequent sections, we first provide a detailed description of these two components and then discuss the training and loss of the model.

\textbf{Image Encoder.}  Given an input image $x$, we exploit pre-trained vision transformers (ViTs) to extract patch embeddings $\mathbf{z} \in \mathbb{R}^{N \times D}$, where $N$ is the number of image patches and $D$ is the embedding dimension.  To capture local features from image patches for the segmentation task, we use the \textit{multi-level feature merging} introduced in COMM~\cite{clip2dino}, which uses both CLIP and DINOv2 to extract the embeddings.  Specifically, given the CLIP model $\text{CLIP}_{visual}$ and an input image $x$, we extract patch embeddings from all transformer blocks $\text{CLIP}_{visual}(x) = [\mathbf{c}^{0}, \mathbf{c}^{1}, \dots, \mathbf{c}^{m-1}]$, where $m$ is the number of transformer blocks.  To align embeddings from different blocks, we apply a linear-layernorm module (LLN)~\cite{clip2dino} to patch embeddings of each block.  The LLN is a layer norm layer followed by a linear layer.  Then, we merge the patch embeddings from different blocks by weighted sum $\overline{\mathbf{c}} = \sum_{i=0}^{m-1} \alpha_i \cdot \text{LLN}(\mathbf{c}^{i})$, where the block scales $\alpha_i$ are learned during training.  The DINOv2 patch embeddings $\overline{\mathbf{d}}$ are also extracted with the same approach.  Note that we only extract patch embeddings from the last $l$ blocks of DINOv2 because the shallow features lead to significant performance degradation~\cite{clip2dino}.  Hence, the DINOv2 patch embeddings are $\overline{\mathbf{d}} = \sum_{i=n-l}^{n-1} \beta_i \cdot \text{LLN}(\mathbf{d}^{i})$.  In order to capture global image features, we also obtain the image embeddings from the $cls$ tokens of CLIP and DINOv2, denoted as $\hat{\mathbf{c}}$ and $\hat{\mathbf{d}}$.  In the end, the output of our image encoder is the patch-wise concatenation of the extracted embeddings as $\mathbf{z} = [\overline{\mathbf{c}}, \hat{\mathbf{c}}, \overline{\mathbf{d}}, \hat{\mathbf{d}}]$.  It is worth mentioning that both CLIP and DINOv2 are frozen during training.  The only trainable parameters in the image encoder are the LLN modules and the block scales (i.e., $\alpha_i$ and $\beta_i$).

\textbf{Segment Embedding Head.}  Given arbitrary segments as prompt, the embedding head aims to extract segment embeddings from the patch embeddings $\mathbf{z}$ and map them to the joint space of vision and language.  Specifically, given a segment $s$, we first calculate the segment’s area within each patch and then normalize it with the patch size to determine the segment’s weight within each patch.  Then, we use these weights to compute the weighted average of the patch embeddings.  Finally, the average embedding is mapped to the vision-language space with a linear layer and serves as the segment embedding $\mathbf{s}$.  Note that we use simple mask pooling and linear projection, which are lightweight and cost-effective to train over a large scale of segment-text pairs.  More sophisticated designs such as prompt encoder~\cite{sam} and cross attention~\cite{flamingo} can also be considered, which we leave for future work.

\textbf{Training and Loss.}  After obtaining the segment embeddings $\mathbf{s}_{0,1,\dots,k-1}$ of a set of segments.  We compute the text embeddings $\mathbf{t}_{0,1,\dots,k-1}$ of the corresponding texts.  Then we use the segment-text contrastive loss to train the model as:
\begin{equation}
\begin{aligned}
L = -\frac{1}{2k} \sum_{i=0}^{k-1} \biggl[ &\log \frac{\exp(\mathbf{s}_i \cdot \mathbf{t}_i / \tau)}{\sum_{j=0}^{k-1} \exp(\mathbf{s}_i \cdot \mathbf{t}_j / \tau)}  \\
&+ \log \frac{\exp(\mathbf{s}_i \cdot \mathbf{t}_i / \tau)}{\sum_{j=0}^{k-1} \exp(\mathbf{s}_j \cdot \mathbf{t}_i / \tau)} \biggr],
\end{aligned}
\end{equation}
where $\tau$ is the temperature parameter that scales the logits.  Note that a segment may correspond to multiple text descriptions in the training data.  At each training iteration, we randomly sample a text description for each segment in the mini-batch to compute the text embedding.

\section{Experiments}

\subsection{Datasets}
\label{sec:datasets}

\begin{table}[t]
  \centering
  \small
  \tablestyle{4pt}{1.0}
  \begin{tabular}{lcccc}
    \toprule
    Dataset & \#pairs & \#pairs w/ NMS & \#expressions \\
    \midrule
    COCO (OVSeg)~\cite{ovseg} & 1.3M & - & 0.3M \\
    \midrule
    COCO & 5.6M & 1.3M & 0.9M \\
    VG & 5.0M & 2.9M & 3.1M \\
    \bottomrule
  \end{tabular}
  \caption{The number of segment-text pairs and unique expressions generated by the proposed data pipeline.}
  \label{tab:data_stat}
\end{table}

\textbf{Training Data.}  We collect training data using the proposed data pipeline from two datasets including COCO~\cite{coco} and Visual Genome (VG)~\cite{visual-genome}.  For COCO, we use all training images with captions, which contain 118k images and 590k captions.  We also use CogVLM-17B~\cite{cogvlm} to generate detailed captions for these training images.  The hyperparameters of CogVLM-17B are set as follows: $temperature=0.8$, $Top \, P=0.4$, and $Top \, K=5$.  The images and the captions are fed into grounding DINO~\cite{GDino} with Swin-T backbone to generate bounding boxes of reference expressions.  The box threshold is set to 0.05, and the NMS threshold is set to 0.7 for grounding DINO.  The bounding boxes are then fed to SAM~\cite{sam} with ViT-H backbone to generate the corresponding segments.  Most of the hyperparameters for SAM are set as the default value, except the IoU threshold and the stability score threshold are both reduced to 0.6 to obtain more segments.  Similar segments are merged using NMS with an IoU threshold of 0.7, and the corresponding expressions are merged into a list.  For VG, we use the human-annotated box-text pairs from the training data directly and convert the boxes into segments using SAM with the same hyperparameter setting as COCO.  The numbers of segment-text pairs and unique expressions are shown in Table~\ref{tab:data_stat}.  Compared with OVSeg~\cite{ovseg}, our data pipeline generates 4 times segment-text pairs and 3 times unique expressions on the COCO dataset, because OVSeg only focuses on nouns.

\textbf{Test Data.}  We evaluate the USE Model on two tasks, including open-vocabulary semantic segmentation and open-vocabulary part segmentation.  For open-vocabulary semantic segmentation, we evaluate our model on ADE20K \cite{ade20k} and Pascal Context \cite{pascal-context} datasets.  ADE20K is a large-scale dataset for scene understanding with 20K training images and 2K validation images.  We use the validation set with two sets of categories for evaluation, one set includes 150 frequently used categories (ADE-150) and the other set contains a full list of 847 categories (ADE-847).  Pascal Context is a dataset for semantic understanding with 4,998 training and 5,105 validation images.  We also use the validation set with two sets of categories for evaluation including one with 59 categories (PC-59) and the other one with 459 categories (PC-459).  For open-vocabulary part segmentation, we perform the experiments on the PartImageNet \cite{part-image-net} dataset, which contains 16,540 training images and 2,957 validation images.  We use the validation set for evaluation, which contains 40 part categories. It is worth mentioning that our model is not trained on any of the training images mentioned above.  Moreover, none of the category names are known before testing.

\subsection{Implementation Details}
\label{sec:implementation_details}

We employ the ViT-L/14 CLIP model pre-trained on 336$\times$336 resolution and the ViT-L/14 distilled DINOv2 model in the image encoder.  For the CLIP model, we collect patch tokens from all transformer blocks and for DINOv2 we only use the patch tokens output from the last 6 transformer blocks.  The embeddings of expressions are generated by the same ViT-L/14 CLIP model with 4 prompt templates including: \textit{a photo of \{\}}, \textit{This is a photo of \{\}}, \textit{There is \{\} in the scene}, and \textit{a photo of \{\} in the scene}.  During training, the input images are augmented with random image resizing with a scaling factor from 0.5 to 2 and random cropping with a size of 560$\times$560.  The USE model is trained on the generated segment-text pairs for 5 epochs with a batch size of 32.  The temperature $\tau$ from the segment-text contrastive loss is set to 30 for all experiments.  We set the initial learning rate to 0.001 and decay it with a polynomial learning rate policy with a power of 0.9.  The AdamW optimizer is used with a weight decay of 0.01.

\subsection{Open-Vocabulary Semantic Segmentation}
\label{sec:semantic_segmentation}

\begin{table*}[t]
  \centering
  \small
  \tablestyle{7pt}{1.0}
  \begin{tabular}{lcccccccc}
    \toprule
    Method & Type & Training Data & VL-Model & ADE-150 & ADE-847 & PC-59 & PC-459 & Average \\
    \midrule
    LSeg+~\cite{lseg} & end2end & COCO & ALIGN EN-B7 & 18.0 & 3.8 & 46.5 & 7.8 & 19.0 \\
    ZegFormer~\cite{zegformer} & end2end & COCO & CLIP ViT-B/16 & 16.4 & - & - & - & - \\
    OpenSeg~\cite{openseg} & end2end & COCO & ALIGN EN-B7 & 28.6 & 8.8 & 48.2 & 12.2 & 24.4 \\
    ODISE~\cite{odise} & end2end & COCO & Stable Diffusion & 29.9 & 11.1 & 57.3 & 14.5 & 28.2 \\
    SAN~\cite{san} & end2end & COCO & CLIP ViT-L/14 & 32.1 & 12.4 & 57.7 & \textbf{15.7} & 29.4 \\
    \midrule
    SimSeg~\cite{simseg} & two-stage & COCO & CLIP ViT-L/14 & 21.7 & 7.1 & 52.2 & 10.2 & 22.8 \\
    OVSeg~\cite{ovseg} & two-stage & COCO & CLIP ViT-L/14 & 29.6 & 9.0 & 55.7 & 12.4 & 26.6 \\
    OVSeg+SAM & two-stage & COCO & CLIP ViT-L/14 & 27.5 & 8.8 & 51.2 & 12.3 & 24.9 \\
    \midrule
    USE+SAM (ours) & two-stage & COCO$\dagger$ & CLIP ViT-L/14 & \textbf{37.0} & \textbf{13.3} & \textbf{57.8} & 14.7 & \textbf{30.7} \\
    \midrule
    \color{gray}USE+SAM (ours) & \color{gray}two-stage & \color{gray}COCO,VG & \color{gray}CLIP ViT-L/14 & \color{gray}37.1 & \color{gray}13.4 & \color{gray}58.0 & \color{gray}15.0 & \color{gray}30.9 \\
    \bottomrule
  \end{tabular}
  \caption{\textbf{Open-vocabulary semantic segmentation benchmarks measured by mIoU.}  Our method outperforms the state-of-the-art two-stage methods by a large margin on all datasets.  Our method also achieves the best average performance compared with all previous methods.  $\dagger$ We use all segment-text pairs from COCO images including the annotations from VG.}
  \label{tab:main}
\end{table*}

We evaluate our method with open-vocabulary semantic segmentation using class-agnostic masks. The class-agnostic masks are generated by prompting SAM with a regular grid of point prompts followed by filtering and merging duplicate masks via NMS.  For each mask, we first obtain its embedding using our model and then compute the similarities between the segment embedding and the text embeddings of the target classes.  Here, we adopt the prompt template used in~\cite{clips4} to generate text embeddings as the class names are mostly nouns.  The similarities are then converted to probabilities with softmax. To generate semantic segmentation maps, we calculate the class prediction of each pixel by aggregating the probabilities of all segments that cover the pixel and taking the class with the highest probability.

We compare the performance of our method with the state-of-the-art open-vocabulary semantic segmentation methods~\cite{lseg, zegformer, openseg, odise, san, simseg, ovseg} on the ADE20K and Pascal Context datasets.  The performance is evaluated with the mean Intersection over Union (mIoU) across all classes.  For methods that were evaluated with different CLIP models~\cite{san, simseg, ovseg}, we use results from the ViT-L/14 CLIP model for comparison.  For other methods, we use the highest performance number of each method for comparison.  We first train our model on the segment-text pairs from the COCO images for fair comparison.  Similar to other two-stage methods~\cite{simseg, ovseg}, we also train an extra model on the COCO ground truth annotations and use predictions from both models to make the final prediction.  Table~\ref{tab:main} shows the benchmarking results on the ADE20K and Pascal Context datasets.  We observe that our method consistently outperforms the state-of-the-art two-stage methods on all datasets by a large margin.  Note that OVSeg's performance declines when using SAM segments, indicating that SAM segments at varying granularities are even more challenging to classify.  Meanwhile, our method archives the best average performance compared with the state-of-the-art end-to-end methods while preserving the flexibility of taking segments at various granularities as prompt.  Extra performance boost can also be observed when training on COCO plus VG images.

\begin{table}[ht]
\centering
\small
\tablestyle{3pt}{1.0}
\begin{tabular}{@{}ccccccc@{}}
\toprule
Method & Datasets  & \multicolumn{1}{c}{All} & \multicolumn{4}{c}{Quadruped} \\ 
\cmidrule(lr){3-3} \cmidrule(lr){4-7}
& & (40) & head & body & foot & tail \\ 
\midrule
USE (ours)& COCO & 6.2 & 8.8 & 2.6 & 2.6 & 18.5 \\
\midrule
VLPart~\cite{vlpart} & Pascal Part  & 4.5 & 17.4 & 0.1 & 0.0 & 2.9 \\
VLPart~\cite{vlpart} & + ImageNet & 5.4 & 23.6 & 3.4 & 0.8 & 1.2 \\
VLPart~\cite{vlpart} & + ImageNet w/ Parts & 7.8 & 35.0 & 15.2 & 3.5 & 8.9 \\
\bottomrule
\end{tabular}
\caption{\textbf{Open-vocabulary part segmentation benchmarks on PartImageNet measured by mAP.} Here, $mAP_{mask}@[0.5,0.95]$ of all 40 parts and Quadruped's parts are presented. Specifically, our method is trained on COCO datasets that do not contain any human-annotated part segments. In contrast, VLPart is first trained on human-annotated part data, Pascal Part. Then, image-level annotations and part-level annotations on ImageNet are added to the training data sequentially.}
\label{tab:part-results}
\end{table}

\subsection{Open-Vocabulary Part Segmentation}
\label{sec:part_segmentation}

In addition to semantic segmentation, the effectiveness of our method is also evaluated with open-vocabulary part segmentation on PartImageNet dataset~\cite{part-image-net}.  To begin with, the class-agnostic masks are generated using SAM.  The similarities between the segment embeddings and text embeddings of the target classes are obtained with the same approach discussed in Section~\ref{sec:semantic_segmentation}.  Because the class-agnostic masks are generated by prompting SAM with uniformly sampled points over the entire image, most of the proposed masks do not contain any object parts.  Instead, it may contain the entire object in the foreground or the objects in the background.  Therefore, we combine the classes of the parts with a list of common background classes to perform classification.  Specifically, we include the 91 COCO stuff classes and 11 super-categories from PartImageNet.  During inference, we only evaluate the masks whose most similar classes are one of the target part categories.

\begin{figure}[t]
  \centering
  \includegraphics[width=\columnwidth]{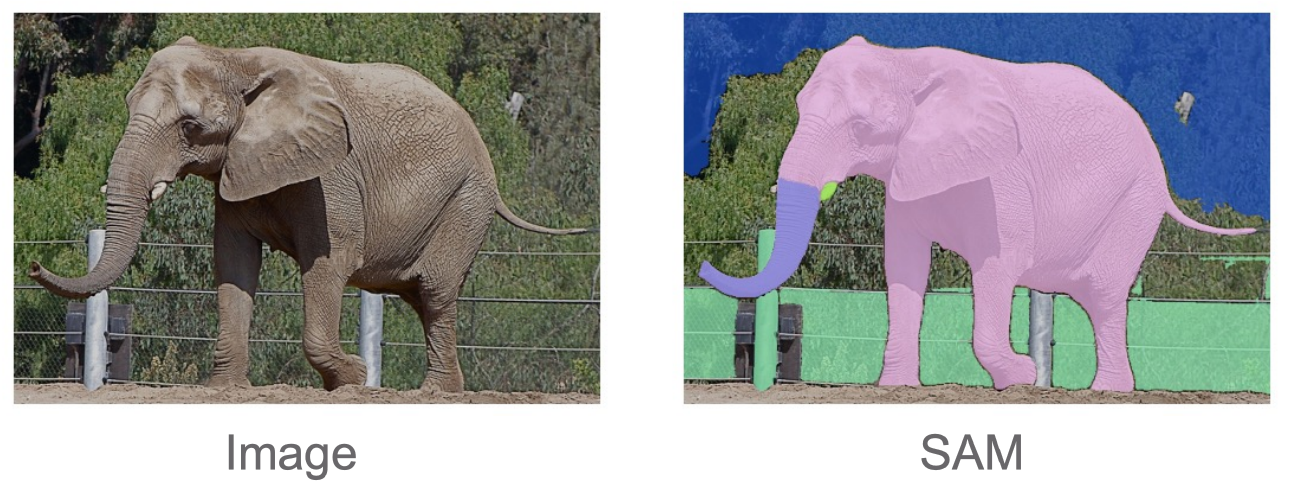}
  \caption{Illustrative example of class-agnostic masks generated by SAM. SAM fails to capture the elephant's head because the boundary lines between the head and the neck are very blurry.}
  \label{fig:sam-elephant}
\end{figure}

To evaluate the performance of our model, we compare it against VLPart~\cite{vlpart} which is specifically designed and trained for open-vocabulary part segmentation.  To assess our open-vocabulary recognition ability on parts against VLPart, we choose to compare with their cross-dataset generalization performance on the PartImageNet dataset, i.e., the VLPart model is trained on datasets other than PartImageNet.  Following VLPart, we adopt $mAP_{mask}@[0.5,0.95]$ as our evaluation metric.  As indicated in Table~\ref{tab:part-results}, our model outperforms VLPart trained on Pascal Part (human-annotated part data) and VLPart trained on Pascal Part + ImageNet over all 40 categories by 1.7 and 0.8, respectively, even though our model was not trained on any human-annotated part data and have not seen any images from ImageNet during training.  Compared with the VLPart trained on Pascal Part + ImageNet + ImageNet w/ Parts (4th row in Table~\ref{tab:part-results}), our performance is slightly worse.  However, it’s important to note that this VLPart model is not under an open-vocabulary setup, as it relies on known target classes of the downstream task and incorporates part segments derived from human annotations.

In terms of the detailed metrics of Quadruped, our model achieved sufficiently high mAP on the body, foot, and especially tail parts, but our model does not perform well in terms of the head.  This is caused by the limitation of SAM because SAM is mostly edge-oriented and thus hardly differentiates two parts if the boundary edges between them are blurry. For example, as shown in Figure~\ref{fig:sam-elephant}, the elephant's trunk has clear edges, whereas the boundary lines between the elephant's head and the elephant's neck are fuzzy and thus can hardly be distinguished by SAM.  It is worth mentioning that, our method is flexible enough to take the class-agnostic masks generated by any image segmentation model.  Hence, segmentation models that are specifically designed for parts can be used to improve our open-vocabulary part segmentation performance.

\subsection{Ablation Study}
\label{sec:ablation_study}

\begin{table}[t]
  \tablestyle{4pt}{1.0}
  \begin{tabular}{lcc}
    \toprule
    Image Encoder & ADE-150 (mIoU) & ADE-847 (mIoU) \\
    \midrule
    CLIP & 30.2 & 10.3 \\
    DINOv2 & 31.9 & 10.2 \\
    CLIP + DINOv2 & \textbf{32.6} & \textbf{11.3} \\
    \bottomrule
  \end{tabular}
  \caption{\textbf{Ablation study} on the choice of the pre-trained backbone. Combining CLIP and DINOv2 gives the best mIoU.}
  \label{tab:backbone}
\end{table}

\begin{table}[t]
  \tablestyle{3pt}{1.0}
  \begin{tabular}{lcc}
    \toprule
    Architecture & ADE-150 (mIoU) & ADE-847 (mIoU) \\
    \midrule
    w/o cls token & 31.4 & 10.0 \\
    w/ cls token & \textbf{31.9} & \textbf{10.2} \\
    \bottomrule
  \end{tabular}
  \caption{\textbf{Ablation study} on architecture design of the image encoder. Only DINOv2 is used in the study.}
  \label{tab:architecture}
\end{table}

We study the choice of the pre-trained backbone on the ADE20K dataset and the open-vocabulary semantic segmentation task.  We train the model on the COCO images and set the crop size of images to 336$\times$336 during training to reduce computation costs.  The evaluation results are shown in Table~\ref{tab:backbone}, which shows that performance gains can be obtained by combining CLIP and DINOv2.

Compared with COMM~\cite{clip2dino}, we propose to concatenate the cls token with the patch tokens when extracting image features for the embedding head.  We study the influence of the cls token on the ADE20K dataset for open-vocabulary semantic segmentation.  The model is trained with the same hyperparameter setting as the previous study.  The performance number with and without using the cls token is shown in Table~\ref{tab:architecture}.  We can see that the mIoU is improved consistently by including the cls token.

\begin{figure}[t]
  \centering
  \includegraphics[width=\columnwidth]{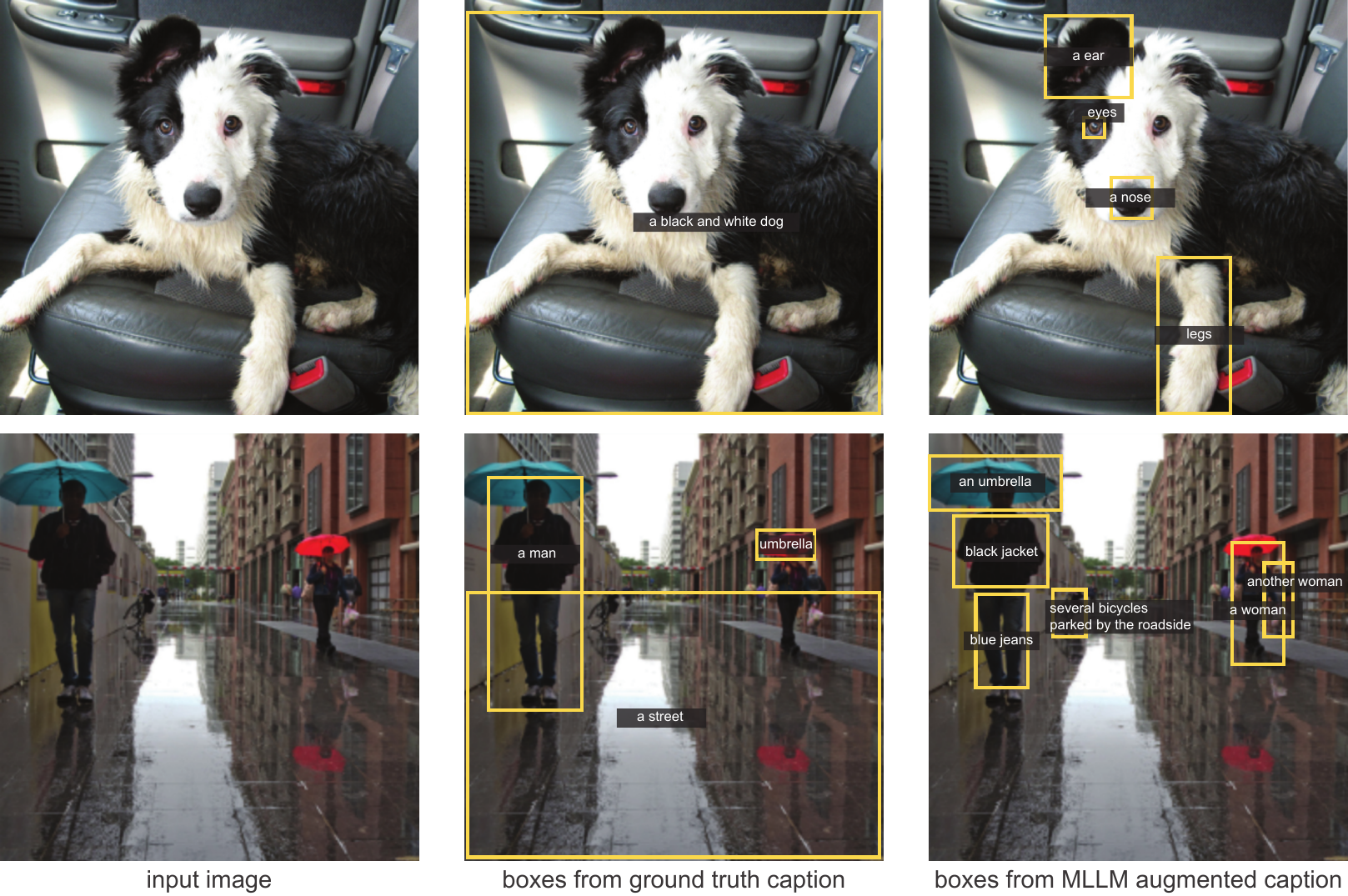}
  \caption{More fine-grained objects and parts can be extracted from MLLM augmented captions compared with ground truth captions.}
  \label{fig:cogvlm}
\end{figure}

We qualitatively compare the objects extracted from ground truth captions and MLLM-augmented captions in Figure~\ref{fig:cogvlm}.  More fine-grained objects and parts can be captured by MLLM-augmented captions compared with ground truth captions.  For example, the eye, nose, ear, and leg of the dog.

\section{Conclusion}

This paper presents the USE framework for open-vocabulary image segmentation.  By integrating a carefully designed data pipeline and a lightweight embedding model, the USE framework effectively classifies image segments in a zero-shot manner without human annotations.  Our approach leverages pre-trained foundation models, optimized for efficiency and scalability.  Extensive experiments demonstrate the superiority of the USE framework over existing methods in semantic and part segmentation.  We hope this work can shed some light on building foundation models for open-vocabulary image segmentation and segment-based representation learning.

{
    \small
    \bibliographystyle{ieeenat_fullname}
    \bibliography{main}

\begin{thebibliography}{40}
\providecommand{\natexlab}[1]{#1}
\providecommand{\url}[1]{\texttt{#1}}
\expandafter\ifx\csname urlstyle\endcsname\relax
  \providecommand{\doi}[1]{doi: #1}\else
  \providecommand{\doi}{doi: \begingroup \urlstyle{rm}\Url}\fi

\bibitem[Alayrac et~al.(2022)Alayrac, Donahue, Luc, Miech, Barr, Hasson, Lenc, Mensch, Millican, Reynolds, Ring, Rutherford, Cabi, Han, Gong, Samangooei, Monteiro, Menick, Borgeaud, Brock, Nematzadeh, Sharifzadeh, Bi\'{n}kowski, Barreira, Vinyals, Zisserman, and Simonyan]{flamingo}
Jean-Baptiste Alayrac, Jeff Donahue, Pauline Luc, Antoine Miech, Iain Barr, Yana Hasson, Karel Lenc, Arthur Mensch, Katherine Millican, Malcolm Reynolds, Roman Ring, Eliza Rutherford, Serkan Cabi, Tengda Han, Zhitao Gong, Sina Samangooei, Marianne Monteiro, Jacob~L Menick, Sebastian Borgeaud, Andy Brock, Aida Nematzadeh, Sahand Sharifzadeh, Miko\l~aj Bi\'{n}kowski, Ricardo Barreira, Oriol Vinyals, Andrew Zisserman, and Kar\'{e}n Simonyan.
\newblock Flamingo: a visual language model for few-shot learning.
\newblock In \emph{NeurIPS}, pages 23716--23736, 2022.

\bibitem[Caesar et~al.(2018)Caesar, Uijlings, and Ferrari]{coco}
Holger Caesar, Jasper Uijlings, and Vittorio Ferrari.
\newblock {COCO-Stuff}: Thing and stuff classes in context.
\newblock In \emph{CVPR}, pages 1209--1218, 2018.

\bibitem[Cao et~al.(2023)Cao, Zhang, Chen, Yang, Du, Zhang, Lu, and Zheng]{cao2023less}
Liangliang Cao, Bowen Zhang, Chen Chen, Yinfei Yang, Xianzhi Du, Wencong Zhang, Zhiyun Lu, and Yantao Zheng.
\newblock Less is more: Removing text-regions improves {CLIP} training efficiency and robustness.
\newblock \emph{arXiv preprint arXiv:2305.05095}, 2023.

\bibitem[Ding et~al.(2022)Ding, Xue, Xia, and Dai]{zegformer}
Jian Ding, Nan Xue, Gui-Song Xia, and Dengxin Dai.
\newblock Decoupling zero-shot semantic segmentation.
\newblock In \emph{CVPR}, pages 11583--11592, 2022.

\bibitem[Ding et~al.(2023)Ding, Wang, and Tu]{ding2022open}
Zheng Ding, Jieke Wang, and Zhuowen Tu.
\newblock Open-vocabulary universal image segmentation with {MaskCLIP}.
\newblock In \emph{ICML}, pages 8090--8102, 2023.

\bibitem[Fan et~al.(2023)Fan, Krishnan, Isola, Katabi, and Tian]{fan2023improving}
Lijie Fan, Dilip Krishnan, Phillip Isola, Dina Katabi, and Yonglong Tian.
\newblock Improving {CLIP} training with language rewrites.
\newblock \emph{arXiv preprint arXiv:2305.20088}, 2023.

\bibitem[Ghiasi et~al.(2022)Ghiasi, Gu, Cui, and Lin]{openseg}
Golnaz Ghiasi, Xiuye Gu, Yin Cui, and Tsung-Yi Lin.
\newblock Scaling open-vocabulary image segmentation with image-level labels.
\newblock In \emph{ECCV}, pages 540--557, 2022.

\bibitem[Gu et~al.(2022)Gu, Lin, Kuo, and Cui]{clip-od}
Xiuye Gu, Tsung-Yi Lin, Weicheng Kuo, and Yin Cui.
\newblock Open-vocabulary object detection via vision and language knowledge distillation.
\newblock In \emph{ICLR}, 2022.

\bibitem[He et~al.(2021)He, Yang, Yang, Kortylewski, Yuan, Chen, Liu, Yang, and Yuille]{part-image-net}
Ju He, Shuo Yang, Shaokang Yang, Adam Kortylewski, Xiaoding Yuan, Jie-Neng Chen, Shuai Liu, Cheng Yang, and Alan Yuille.
\newblock {PartImageNet}: A large, high-quality dataset of parts.
\newblock \emph{arXiv preprint arXiv:2112.00933}, 2021.

\bibitem[He et~al.(2023)He, Jamonnak, Gou, and Ren]{clips4}
Wenbin He, Suphanut Jamonnak, Liang Gou, and Liu Ren.
\newblock {CLIP-S4}: Language-guided self-supervised semantic segmentation.
\newblock In \emph{CVPR}, pages 11207--11216, 2023.

\bibitem[Honnibal et~al.(2020)Honnibal, Montani, Van~Landeghem, and Boyd]{spacy}
Matthew Honnibal, Ines Montani, Sofie Van~Landeghem, and Adriane Boyd.
\newblock {spaCy}: Industrial-strength natural language processing in python.
\newblock 2020.

\bibitem[Huynh et~al.(2022)Huynh, Kuen, Lin, Gu, and Elhamifar]{huynh2022open}
Dat Huynh, Jason Kuen, Zhe Lin, Jiuxiang Gu, and Ehsan Elhamifar.
\newblock Open-vocabulary instance segmentation via robust cross-modal pseudo-labeling.
\newblock In \emph{CVPR}, pages 7020--7031, 2022.

\bibitem[Jiang et~al.(2023)Jiang, Liu, Liu, Zhang, Li, Xiong, and Tian]{clip2dino}
Dongsheng Jiang, Yuchen Liu, Songlin Liu, Xiaopeng Zhang, Jin Li, Hongkai Xiong, and Qi Tian.
\newblock From {CLIP} to {DINO}: Visual encoders shout in multi-modal large language models.
\newblock \emph{arXiv preprint arXiv:2310.08825}, 2023.

\bibitem[Kim et~al.(2023)Kim, Angelova, and Kuo]{CoDet}
Dahun Kim, Anelia Angelova, and Weicheng Kuo.
\newblock Region-aware pretraining for open-vocabulary object detection with vision transformers.
\newblock In \emph{CVPR}, pages 11144--11154, 2023.

\bibitem[Kirillov et~al.(2023)Kirillov, Mintun, Ravi, Mao, Rolland, Gustafson, Xiao, Whitehead, Berg, Lo, Dollar, and Girshick]{sam}
Alexander Kirillov, Eric Mintun, Nikhila Ravi, Hanzi Mao, Chloe Rolland, Laura Gustafson, Tete Xiao, Spencer Whitehead, Alexander~C. Berg, Wan-Yen Lo, Piotr Dollar, and Ross Girshick.
\newblock Segment anything.
\newblock In \emph{ICCV}, pages 4015--4026, 2023.

\bibitem[Krishna et~al.(2017)Krishna, Zhu, Groth, Johnson, Hata, Kravitz, Chen, Kalantidis, Li, Shamma, et~al.]{visual-genome}
Ranjay Krishna, Yuke Zhu, Oliver Groth, Justin Johnson, Kenji Hata, Joshua Kravitz, Stephanie Chen, Yannis Kalantidis, Li-Jia Li, David~A Shamma, et~al.
\newblock Visual genome: Connecting language and vision using crowdsourced dense image annotations.
\newblock \emph{IJCV}, 123\penalty0 (1):\penalty0 32--73, 2017.

\bibitem[Krizhevsky and Hinton(2009)]{cifar100}
Alex Krizhevsky and Geoffrey Hinton.
\newblock Learning multiple layers of features from tiny images.
\newblock \emph{Master's thesis, Department of Computer Science, University of Toronto}, 2009.

\bibitem[Lai et~al.(2023)Lai, Zhang, Wu, Bai, Timofeev, Du, Gan, Shan, Chuah, Yang, and Cao]{lai2023scarcity}
Zhengfeng Lai, Haotian Zhang, Wentao Wu, Haoping Bai, Aleksei Timofeev, Xianzhi Du, Zhe Gan, Jiulong Shan, Chen-Nee Chuah, Yinfei Yang, and Meng Cao.
\newblock From scarcity to efficiency: Improving {CLIP} training via visual-enriched captions.
\newblock \emph{arXiv preprint arXiv:2310.07699}, 2023.

\bibitem[Li et~al.(2022)Li, Weinberger, Belongie, Koltun, and Ranftl]{lseg}
Boyi Li, Kilian~Q Weinberger, Serge Belongie, Vladlen Koltun, and Ren{\'e} Ranftl.
\newblock Language-driven semantic segmentation.
\newblock In \emph{ICLR}, 2022.

\bibitem[Liang et~al.(2023)Liang, Wu, Dai, Li, Zhao, Zhang, Zhang, Vajda, and Marculescu]{ovseg}
Feng Liang, Bichen Wu, Xiaoliang Dai, Kunpeng Li, Yinan Zhao, Hang Zhang, Peizhao Zhang, Peter Vajda, and Diana Marculescu.
\newblock Open-vocabulary semantic segmentation with mask-adapted {CLIP}.
\newblock In \emph{CVPR}, pages 7061--7070, 2023.

\bibitem[Liu et~al.(2023{\natexlab{a}})Liu, Li, Wu, and Lee]{llava}
Haotian Liu, Chunyuan Li, Qingyang Wu, and Yong~Jae Lee.
\newblock Visual instruction tuning.
\newblock \emph{arXiv preprint arXiv:2304.08485}, 2023{\natexlab{a}}.

\bibitem[Liu et~al.(2023{\natexlab{b}})Liu, Zeng, Ren, Li, Zhang, Yang, Li, Yang, Su, Zhu, and Zhang]{GDino}
Shilong Liu, Zhaoyang Zeng, Tianhe Ren, Feng Li, Hao Zhang, Jie Yang, Chunyuan Li, Jianwei Yang, Hang Su, Jun Zhu, and Lei Zhang.
\newblock {Grounding DINO}: Marrying {DINO} with grounded pre-training for open-set object detection.
\newblock \emph{arXiv preprint arXiv:2303.05499}, 2023{\natexlab{b}}.

\bibitem[Maini et~al.(2023)Maini, Goyal, Lipton, Kolter, and Raghunathan]{maini2023t}
Pratyush Maini, Sachin Goyal, Zachary~C. Lipton, J.~Zico Kolter, and Aditi Raghunathan.
\newblock {T-MARS}: Improving visual representations by circumventing text feature learning.
\newblock \emph{arXiv preprint arXiv:2307.03132}, 2023.

\bibitem[Mokady et~al.(2021)Mokady, Hertz, and Bermano]{clipcap}
Ron Mokady, Amir Hertz, and Amit~H. Bermano.
\newblock {ClipCap}: {CLIP} prefix for image captioning.
\newblock \emph{arXiv preprint arXiv:2111.09734}, 2021.

\bibitem[Mottaghi et~al.(2014)Mottaghi, Chen, Liu, Cho, Lee, Fidler, Urtasun, and Yuille]{pascal-context}
Roozbeh Mottaghi, Xianjie Chen, Xiaobai Liu, Nam-Gyu Cho, Seong-Whan Lee, Sanja Fidler, Raquel Urtasun, and Alan Yuille.
\newblock The role of context for object detection and semantic segmentation in the wild.
\newblock In \emph{CVPR}, pages 891--898, 2014.

\bibitem[Oquab et~al.(2023)Oquab, Darcet, Moutakanni, Vo, Szafraniec, Khalidov, Fernandez, Haziza, Massa, El-Nouby, Howes, Huang, Xu, Sharma, Li, Galuba, Rabbat, Assran, Ballas, Synnaeve, Misra, Jegou, Mairal, Labatut, Joulin, and Bojanowski]{dinov2}
Maxime Oquab, Timothée Darcet, Theo Moutakanni, Huy~V. Vo, Marc Szafraniec, Vasil Khalidov, Pierre Fernandez, Daniel Haziza, Francisco Massa, Alaaeldin El-Nouby, Russell Howes, Po-Yao Huang, Hu Xu, Vasu Sharma, Shang-Wen Li, Wojciech Galuba, Mike Rabbat, Mido Assran, Nicolas Ballas, Gabriel Synnaeve, Ishan Misra, Herve Jegou, Julien Mairal, Patrick Labatut, Armand Joulin, and Piotr Bojanowski.
\newblock {DINOv2}: Learning robust visual features without supervision.
\newblock \emph{arXiv preprint arXiv:2304.07193}, 2023.

\bibitem[Ordonez et~al.(2011)Ordonez, Kulkarni, and Berg]{sbu}
Vicente Ordonez, Girish Kulkarni, and Tamara~L Berg.
\newblock {Im2Text}: Describing images using 1 million captioned photographs.
\newblock In \emph{NeurIPS}, pages 1143--1151, 2011.

\bibitem[Peng et~al.(2023)Peng, Wang, Dong, Hao, Huang, Ma, and Wei]{kosmos2}
Zhiliang Peng, Wenhui Wang, Li Dong, Yaru Hao, Shaohan Huang, Shuming Ma, and Furu Wei.
\newblock Kosmos-2: Grounding multimodal large language models to the world.
\newblock \emph{arXiv preprint arXiv:2306.14824}, 2023.

\bibitem[Radford et~al.(2021)Radford, Kim, Hallacy, Ramesh, Goh, Agarwal, Sastry, Askell, Mishkin, Clark, Krueger, and Sutskever]{clip}
Alec Radford, Jong~Wook Kim, Chris Hallacy, Aditya Ramesh, Gabriel Goh, Sandhini Agarwal, Girish Sastry, Amanda Askell, Pamela Mishkin, Jack Clark, Gretchen Krueger, and Ilya Sutskever.
\newblock Learning transferable visual models from natural language supervision.
\newblock In \emph{ICML}, pages 8748--8763, 2021.

\bibitem[Sharma et~al.(2018)Sharma, Ding, Goodman, and Soricut]{cc3m}
Piyush Sharma, Nan Ding, Sebastian Goodman, and Radu Soricut.
\newblock Conceptual captions: A cleaned, hypernymed, image alt-text dataset for automatic image captioning.
\newblock In \emph{ACL}, pages 2556--2565, 2018.

\bibitem[Sun et~al.(2023)Sun, Chen, Zhu, Xiao, Luo, Xie, and Yan]{vlpart}
Peize Sun, Shoufa Chen, Chenchen Zhu, Fanyi Xiao, Ping Luo, Saining Xie, and Zhicheng Yan.
\newblock Going denser with open-vocabulary part segmentation.
\newblock \emph{arXiv preprint arXiv:2305.11173}, 2023.

\bibitem[Wang et~al.(2023)Wang, Lv, Yu, Hong, Qi, Wang, Ji, Yang, Zhao, Song, Xu, Xu, Li, Dong, Ding, and Tang]{cogvlm}
Weihan Wang, Qingsong Lv, Wenmeng Yu, Wenyi Hong, Ji Qi, Yan Wang, Junhui Ji, Zhuoyi Yang, Lei Zhao, Xixuan Song, Jiazheng Xu, Bin Xu, Juanzi Li, Yuxiao Dong, Ming Ding, and Jie Tang.
\newblock {CogVLM}: Visual expert for pretrained language models.
\newblock \emph{arXiv preprint arXiv:2311.03079}, 2023.

\bibitem[Xu et~al.(2022{\natexlab{a}})Xu, De~Mello, Liu, Byeon, Breuel, Kautz, and Wang]{groupvit}
Jiarui Xu, Shalini De~Mello, Sifei Liu, Wonmin Byeon, Thomas Breuel, Jan Kautz, and Xiaolong Wang.
\newblock {GroupViT}: Semantic segmentation emerges from text supervision.
\newblock In \emph{CVPR}, pages 18134--18144, 2022{\natexlab{a}}.

\bibitem[Xu et~al.(2023{\natexlab{a}})Xu, Liu, Vahdat, Byeon, Wang, and De~Mello]{odise}
Jiarui Xu, Sifei Liu, Arash Vahdat, Wonmin Byeon, Xiaolong Wang, and Shalini De~Mello.
\newblock Open-vocabulary panoptic segmentation with text-to-image diffusion models.
\newblock In \emph{CVPR}, pages 2955--2966, 2023{\natexlab{a}}.

\bibitem[Xu et~al.(2022{\natexlab{b}})Xu, Zhang, Wei, Lin, Cao, Hu, and Bai]{simseg}
Mengde Xu, Zheng Zhang, Fangyun Wei, Yutong Lin, Yue Cao, Han Hu, and Xiang Bai.
\newblock A simple baseline for open-vocabulary semantic segmentation with pre-trained vision-language model.
\newblock In \emph{ECCV}, pages 736--753, 2022{\natexlab{b}}.

\bibitem[Xu et~al.(2023{\natexlab{b}})Xu, Zhang, Wei, Hu, and Bai]{san}
Mengde Xu, Zheng Zhang, Fangyun Wei, Han Hu, and Xiang Bai.
\newblock Side adapter network for open-vocabulary semantic segmentation.
\newblock In \emph{CVPR}, pages 2945--2954, 2023{\natexlab{b}}.

\bibitem[Zhong et~al.(2022)Zhong, Yang, Zhang, Li, Codella, Li, Zhou, Dai, Yuan, Li, and Gao]{regionclip}
Yiwu Zhong, Jianwei Yang, Pengchuan Zhang, Chunyuan Li, Noel Codella, Liunian~Harold Li, Luowei Zhou, Xiyang Dai, Lu Yuan, Yin Li, and Jianfeng Gao.
\newblock {RegionCLIP}: Region-based language-image pretraining.
\newblock In \emph{CVPR}, pages 16793--16803, 2022.

\bibitem[Zhou et~al.(2017)Zhou, Zhao, Puig, Fidler, Barriuso, and Torralba]{ade20k}
Bolei Zhou, Hang Zhao, Xavier Puig, Sanja Fidler, Adela Barriuso, and Antonio Torralba.
\newblock Scene parsing through {ADE20K} dataset.
\newblock In \emph{CVPR}, pages 633--641, 2017.

\bibitem[Zhu et~al.(2023)Zhu, Chen, Haydarov, Shen, Zhang, and Elhoseiny]{zhu2023chatgpt}
Deyao Zhu, Jun Chen, Kilichbek Haydarov, Xiaoqian Shen, Wenxuan Zhang, and Mohamed Elhoseiny.
\newblock {ChatGPT} asks, {BLIP}-2 answers: Automatic questioning towards enriched visual descriptions.
\newblock \emph{arXiv preprint arXiv:2303.06594}, 2023.

\bibitem[Zou et~al.(2023)Zou, Dou, Yang, Gan, Li, Li, Dai, Behl, Wang, Yuan, Peng, Wang, Lee, and Gao]{x-decoder}
Xueyan Zou, Zi-Yi Dou, Jianwei Yang, Zhe Gan, Linjie Li, Chunyuan Li, Xiyang Dai, Harkirat Behl, Jianfeng Wang, Lu Yuan, Nanyun Peng, Lijuan Wang, Yong~Jae Lee, and Jianfeng Gao.
\newblock Generalized decoding for pixel, image, and language.
\newblock In \emph{CVPR}, pages 15116--15127, 2023.

\end{thebibliography}
}


\end{document}


\maketitle
\appendix

\section{Supplementary}
\label{sec:supplementary}

In this supplementary, we include:
\begin{itemize}
  \item visualization of curated data (\ref{sec:vis_data})
  \item additional ablation study (\ref{sec:ablation_study_add})
  \item additional quantitative results on open-vocabulary semantic segmentation (\ref{sec:quantitative_add})
  \item qualitative comparison of open-vocabulary semantic segmentation methods (\ref{sec:qualitative})
  \item visualization of text-based query (\ref{sec:vis_query})
\end{itemize}

\subsection{Examples of Curated Data}
\label{sec:vis_data}

Figure~\ref{fig:data} shows 4 example images with generated segment text pairs.  The captions of the 4 images are listed as follows:
\begin{itemize}
  \item Figure~\ref{fig:data}a:
    \begin{itemize}
        \item A toy dinosaur standing on a sink next to a running faucet. (human-annotated)
        \item a faucet running next to a dinosuar holding a toothbrush (human-annotated) 
        \item A toy lizard with a toothbrush in it's mouth standing next to a running water faucet in a bathroom. (human-annotated)
        \item A fake toy dinousure has a green tooth brush in its mouth (human-annotated)
        \item a sink with running water a mirror and a Godzilla toothbrush holder (human-annotated)
        \item \textit{In the picture, there is an orange dinosaur toy in front of a white sink. The dinosaur figurine is standing on its hind legs, with a blue toothbrush in its mouth. Above it is a mirror, and behind the dinosaur toy, there is a silver metal faucet. Next to the sink, there is a round hole. (MLLM-generated)}
    \end{itemize}
\begin{figure}[t]
  \centering
  \includegraphics[width=\columnwidth]{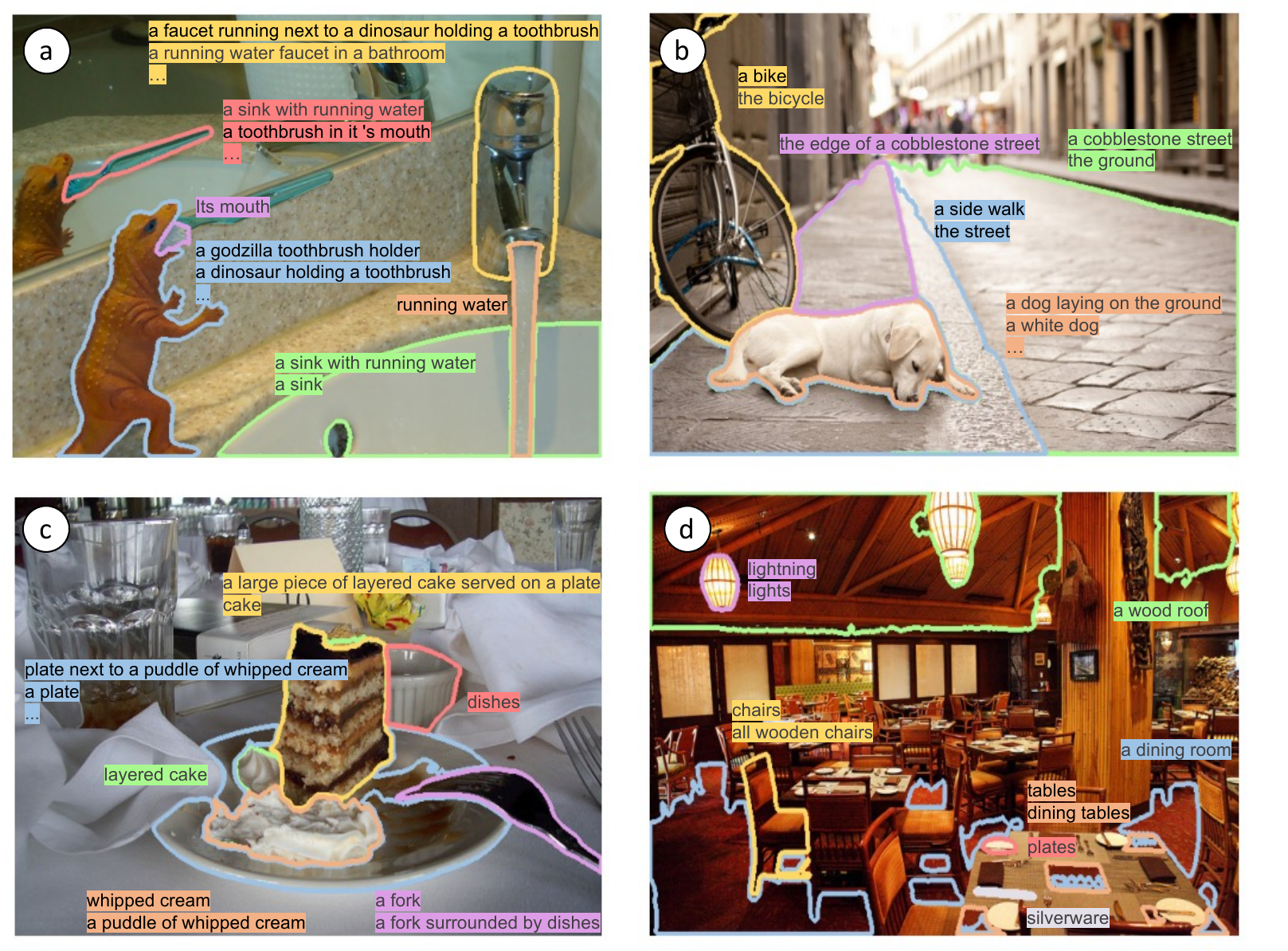}
  \caption{Examples of the curated segment text pairs.}
  \label{fig:data}
\end{figure}
\begin{table*}[t]
  \centering
  \small
  \tablestyle{8pt}{1.0}
  \begin{tabular}{lcccccccc}
    \toprule
    Method & Type & Training Data & VL-Model & ADE-150 & ADE-847 & PC-59 & PC-459 & Average \\
    \midrule
    USE & two-stage & COCO w/o GT mask & CLIP ViT-L/14 & 35.2 & 13.2 & 54.6 & 13.6 & 29.2 \\
    USE & two-stage & COCO & CLIP ViT-L/14 & 37.0 & 13.3 & 57.8 & 14.7 & 30.7 \\
    \bottomrule
  \end{tabular}
  \caption{Influence of combining models trained on different data sources.  Performance is evaluated using mIoU.}
  \label{tab:combine}
\end{table*}
\begin{table*}[t]
  \centering
  \small
  \tablestyle{8pt}{1.0}
  \begin{tabular}{lcccccc}
    \toprule
    Method & Training Data & Metric & ADE-150 & ADE-847 & PC-59 & Average \\
    \midrule
    SAN~\cite{san} & COCO & pAcc & 71.0 & 57.2 & 78.8 & 69.0 \\
    USE & COCO & pAcc & 70.7 & 58.7 & 77.4 & 68.9 \\
    \midrule
    SAN~\cite{san} & COCO & mAcc & 50.7 & 25.3 & 77.5 & 51.1 \\
    USE & COCO & mAcc & 58.4 & 29.8 & 77.3 & 55.1 \\
    \bottomrule
  \end{tabular}
  \caption{Open-vocabulary semantic segmentation benchmarks measured by pAcc and mAcc.}
  \label{tab:acc}
\end{table*}

  \item Figure~\ref{fig:data}b:
    \begin{itemize}
        \item A picture of a dog laying on the ground. (human-annotated)
        \item Dog snoozing by a bike on the edge of a cobblestone street (human-annotated) 
        \item The white dog lays next to the bicycle on the sidewalk. (human-annotated)
        \item a white dog is sleeping on a street and a bicycle (human-annotated)
        \item A puppy rests on the street next to a bicycle. (human-annotated)
        \item \textit{A small dog is lying on the sidewalk, sleeping. There is a bicycle leaning against the wall with black spokes and a blue basket. On both sides of the street are tall buildings. In front of the building on the right, there are several people walking or standing. (MLLM-generated)}
    \end{itemize}
  \item Figure~\ref{fig:data}c:
    \begin{itemize}
        \item A slice of cake on a plate next to a puddle of whipped cream. (human-annotated)
        \item A half-eaten piece of cake sits on a plate on a cluttered table. (human-annotated) 
        \item Multiple layer cake on plate, half eaten with a fork. (human-annotated)
        \item A nicely set table with a large piece of layered cake served on a plate. (human-annotated)
        \item a piece of cake on a plate with a fork surrounded by dishes (human-annotated)
        \item \textit{The image shows a half-eaten, multi-layered cake in a large plate placed on a cluttered dining table. A fork is resting next to the plate, indicating that someone has already started enjoying their dessert. In addition to the cake and utensil, there are several other items scattered across the table: a bowl, multiple cups, two books, and some bottles. The scene appears quite busy with various objects occupying space on the table, suggesting it might be during a gathering or celebration where people have come together for food and drinks. (MLLM-generated)}
    \end{itemize}
  \item Figure~\ref{fig:data}d:
    \begin{itemize}
        \item A dining room filled with tables and chairs and lighting. (human-annotated)
        \item A restaurant with tables, chairs, lights, plates and silverware. (human-annotated) 
        \item A restaurant has tables and chairs and a wood roof. (human-annotated)
        \item A large very large restaurant with all wooden chair and tables. (human-annotated)
        \item A room filled with dining tables and chairs. (human-annotated)
        \item \textit{This picture depicts a dining room with wooden floors and walls, and many tables and chairs arranged in an orderly manner. There are plate, fork, and spoon on each table, as well as napkins. In addition, there is a yellow lamp hanging above the table, and the ceiling is decorated with bamboo and wood structures. On the far right of the photo, there is a pillar with an ancient vase placed on it, and next to it is a wall covered with black curtains. (MLLM-generated)}
    \end{itemize}
\end{itemize}

The MLLM-generated captions capture fine-grained details of the images.  The generated segments with respect to the phrases in the captions are highly relevant, though there are still a few misaligned segments and phrases.

\begin{figure}[t]
  \centering
  \includegraphics[width=\columnwidth]{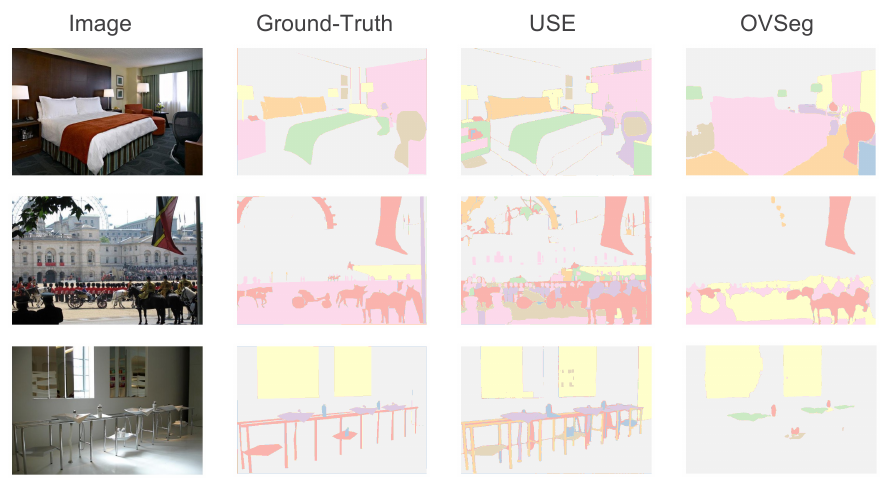}
  \caption{Visual results of open-vocabulary semantic segmentation on ADE20K validation set.  Compared with OVSeg, our method predicts segmentation classes more accurately.}
  \label{fig:pred}
\end{figure}

\subsection{Ablation Study}
\label{sec:ablation_study_add}

Table~\ref{tab:combine} demonstrates the influence of combining models trained on different data sources, including human-annotated and curated masks.  Same as OVSeg~\cite{ovseg}, different models' predictions are combined with different weights.  We set the weight of the model trained on curated masks as 1.  The weight of the model trained on human-annotated masks is set to 0.7 for the Pascal Context dataset with 59 categories and 0.25 for other datasets.  We can see that by using human-annotated masks from the COCO dataset, the performance improves especially for the Pascal Context dataset with 59 categories.  The reason is that the 59 categories of the Pascal Context dataset are similar to the human annotations in the COCO dataset~\cite{san}.  For datasets that contain a large number of categories, the influence of human annotations decreases, for example, the ADE20K dataset with 847 categories.

\subsection{Quantitative Results on Open-Vocabulary Semantic Segmentation}
\label{sec:quantitative_add}

\begin{figure*}[t]
  \centering
  \includegraphics[width=\linewidth]{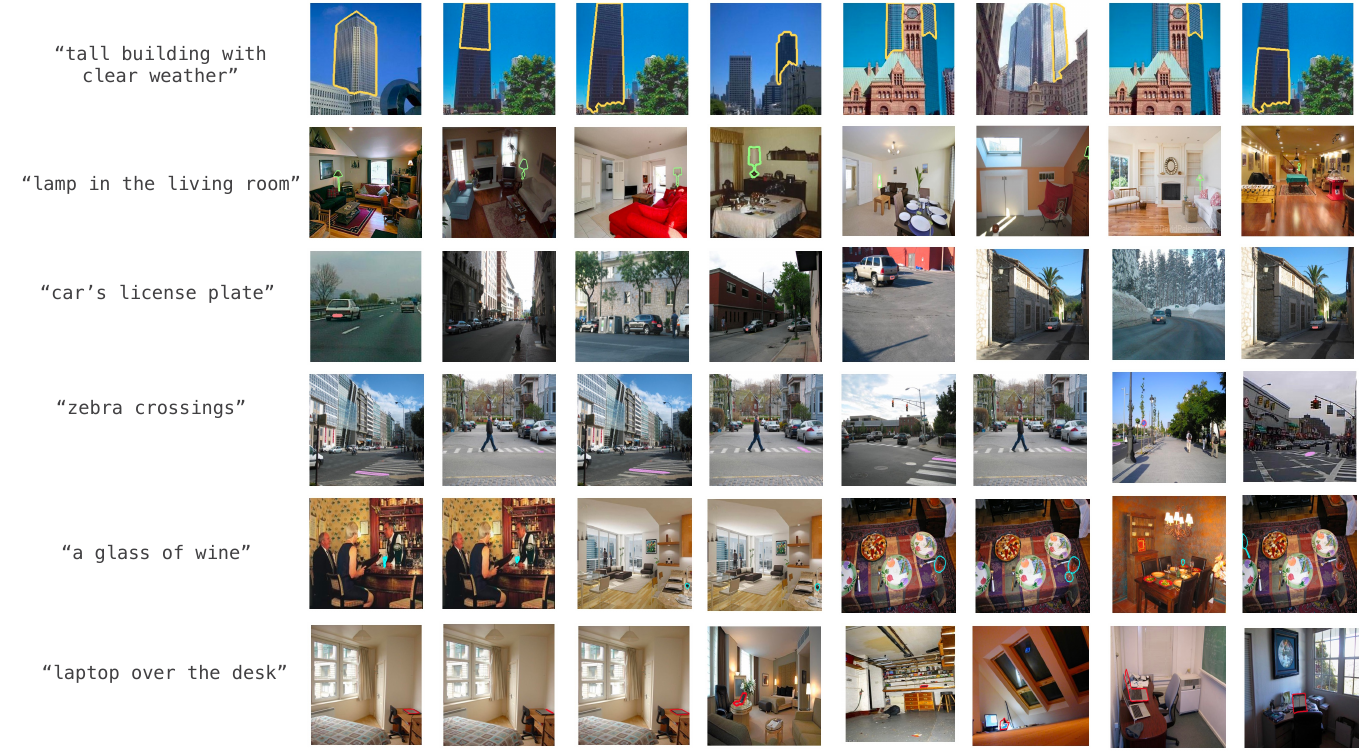}
  \caption{Image segments retrieved for various phrases.  The generated segment embeddings accurately capture the semantic meaning of the segments at various granularities.}
  \label{fig:query}
\end{figure*}

Table~\ref{tab:acc} shows additional quantitative results for open-vocabulary semantic segmentation on pixel accuracy (pAcc) and mean pixel accuracy (mAcc).  Compared with SAN~\cite{san}, our method archives comparable performance in terms of pAcc and obtains much higher mAcc, which indicates that our method performs better on small but challenging objects.

\subsection{Visual Results on Open-Vocabulary Semantic Segmentation}
\label{sec:qualitative}

Figure~\ref{fig:pred} shows visual results of open-vocabulary semantic segmentation on the ADE20K validation set with 150 categories.  Compared with OVSeg, our segment classification results are more accurate and precise.

\subsection{Visual Results on Text-based Query}
\label{sec:vis_query}

Figure~\ref{fig:query} shows the image segments retrieved from the ADE20K validation data for various phrases.  We first obtain the segment embeddings for SAM-generated segments and the text embeddings of the phrases.  Then we compute the similarity between the text and segment embeddings.  We retrieve and visualize the top similar segments for each phrase.  We observe that the segment embeddings can capture the semantic meaning of segments at various granularities, such as the fine-grained concept of ``\textit{car’s license plate}".  Meanwhile, the segment embeddings also capture the context information such as ``\textit{clear weather}", ``\textit{living room}", and ``\textit{the desk}".

{
    \small
    \bibliographystyle{ieeenat_fullname}
    \bibliography{main}
}
